\documentclass[letterpaper, 10 pt, conference]{ieeeconf}  

\IEEEoverridecommandlockouts                              

\usepackage{cite, epsfig, mathptmx, times, amssymb, amsmath, color}
\usepackage{graphicx,booktabs,soul,color,graphics,epsfig,mathptmx,moreverb,ifpdf,cancel,framed,dblfloatfix}
\usepackage{gensymb,graphicx,url, multirow}
\usepackage{pbox}
\usepackage{adjustbox}
\usepackage[linesnumbered,ruled,vlined]{algorithm2e}

\newtheorem{assumption}{Assumption}
\newtheorem{problem}{Problem}
\usepackage[x11names,dvipsnames,table]{xcolor} 
\usepackage{colortbl}
\usepackage[caption=false]{subfig}
\usepackage{floatrow}
\usepackage{float}
\usepackage{caption}
\usepackage{balance}
\usepackage{xcolor}
\usepackage{amsmath,amssymb,amsfonts}
\usepackage{textcomp}
\usepackage{hyperref}

\DeclareMathAlphabet{\mathpzc}{OT1}{pzc}{m}{it}

\DeclareMathOperator*{\argmin}{arg\,min}

\def\BibTeX{{\rm B\kern-.05em{\sc i\kern-.025em b}\kern-.08em
		T\kern-.1667em\lower.7ex\hbox{E}\kern-.125emX}}
\title{\LARGE \bf
Aerial Robot Control in Close Proximity to Ceiling: \\ A Force Estimation-based Nonlinear MPC}

\author{Basaran Bahadir Kocer$^{1,2}$, Mehmet Efe Tiryaki$^{3}$, Mahardhika Pratama$^{4}$, Tegoeh Tjahjowidodo$^{1}$, \\ and Gerald Gim Lee Seet$^{1}$  
\thanks{$^{1}$School of Mechanical and Aerospace Engineering, Nanyang Technological University, 50 Nanyang Avenue, Singapore. Email: koce0001@e.ntu.edu.sg, ttegoeh@ntu.edu.sg, mglseet@ntu.edu.sg.}%
\thanks{$^{2}$Energy Research Institute @ NTU, Nanyang Technological University, 1 CleanTech Loop, Singapore.}%
\thanks{$^{3}$Physical Intelligence Department, Max Planck Institute for Intelligent Systems, Stuttgart 70569, Germany. Email: tiryaki@is.mpg.de.}%
\thanks{$^{4}$School of Computer Science and Engineering, Nanyang Technological University, 50 Nanyang Avenue, Singapore. Email: mpratama@ntu.edu.sg.}%
}

\begin{document}

\maketitle
\thispagestyle{empty}
\pagestyle{empty}

\begin{abstract}
Being motivated by ceiling inspection applications via unmanned aerial vehicles (UAVs) which require close proximity flight to surfaces, a systematic control approach enabling safe and accurate close proximity flight is proposed in this work. There are two main challenges for close proximity flights: (i) the trust characteristics varies drastically for the different distance from the ceiling which results in a complex nonlinear dynamics; (ii) the system needs to consider physical and environmental constraints to safely fly in close proximity. To address these challenges, a novel framework consisting of a constrained optimization-based force estimation and an optimization-based nonlinear controller is proposed. Experimental results illustrate that the performance of the proposed control approach can stabilize UAV down to 1 cm distance to the ceiling. Furthermore, we report that the UAV consumes up to 12.5\% less power when it is operated 1 cm distance to ceiling, which is promising potential for more battery-efficient inspection flights.
%
\end{abstract}
%

\section{Introduction}
\label{sec:introduction}
In recent years, in-site UAV inspection has gained momentum as an application area in robotic research \cite{bircher2018receding}. Typical inspection application requires a robot to achieve accurate motions in close proximity to the environment for long periods of measurement \cite{kocer2019inspection}. This is a challenging control task for conventional controllers such as PID controller, because of the strong cross-coupling between the UAV and surroundings. Although nonlinear controllers for such operating conditions are extensively studied for ground-effects, to our knowledge, there is not any controller systematically using force estimation in predictive control framework to achieve accurate control in close proximity to the ground as well as the ceiling.  

In this study, the system is modeled as (i) a baseline model, which is composed of second-order translational dynamics of UAV system; and (ii) an additive model, which summarizes the interactions of the UAV with its surrounding. The additive model is constructed using lumped external forces. In our proposed approach, these external forces are estimated using nonlinear moving horizon estimation (NMHE) and fed into the nonlinear model predictive controller (NMPC) to fully capture the effect of interaction on the system. Later, the proposed controller is tested in close proximity to the ceiling. In summary, the following novelties are proposed in this work:
\begin{itemize}	
	\item For the first time, an optimization-based framework consisting of force estimation-based nonlinear model predictive controller is investigated for the ceiling effect. 
	\item The power consumption of the aerial robot is analyzed for the proposed system in close proximity to the ceiling.
\end{itemize}

Leveraging the above-listed key findings of this study, it is possible to prolong the flight duration as well as the battery life. Interested readers may refer to the work on the reduced current effect on the battery life for UAVs in \cite{kocer2018uav}.

\begin{figure}[t]
	\centering
	\includegraphics[width=\columnwidth]{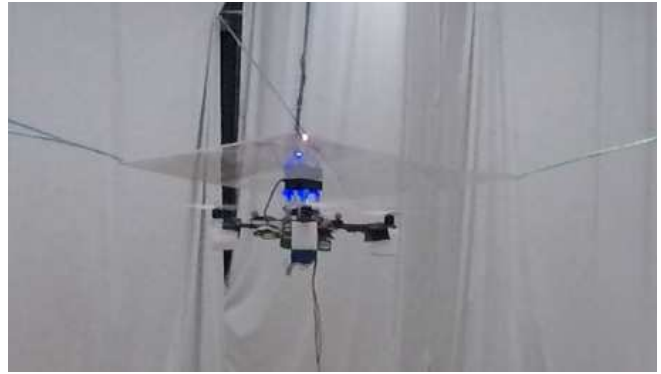}
	\caption{Aerial robot flies below the ceiling. The presented experiments are available at \url{https://youtu.be/AtlyNwIE3kY}}
	\label{fig:exp}
\end{figure}

\section{Related Work}
\label{sec:related_work}
The ground effect observation is reported in landing problems, e.g. in \cite{huber2013first}, and addressed in some implementations. As it can be visualized, the ground effect stands as a repulsive force which still provides a safety margin in a normal flight. For the modeling of the ground effect, the interested reader can refer to \cite{conyers2018empirical,bernard2018ground}. On the other hand, the ceiling effect creates an attractive force which pulls the aerial robot through the surroundings. Without a proper control strategy, the flight of the aerial robot might end up with a crash. Moreover, the near wall effect is numerically investigated in \cite{robinson2014computational}, and experimentally tested in \cite{tavora2017feasibility}; therefore, some control techniques are suggested. It is noted that this effect is milder as compared to the ground and ceiling effects.

The numerical and experimental evaluation of the ceiling effect received less attention than that of ground and near-wall effects so far. However, recent attempts intent to explore its modeling such as \cite{hsiao_2018,conyers2018empiricalc}. In terms of the control, from a passive point of view, it aims to avoid when a downwash effect (aerodynamic effect on the system surrounded by air flow) on the system is detected \cite{yeo2015onboard,yeo2016downwash}. In this context, the path planner generates an alternative trajectory to minimize the downwash effect. A counterintuitive approach to those of avoiding ones is analyzed in \cite{sanchez2017experimental} and the experimental results are presented in \cite{sanchez2017multirotor}, where the rotors are covered by a protective case to be in contact with the ceiling. Similarly, a system is designed to be hooked on some anchor points on the ceiling in \cite{delamare2018toward}. However, its active compensation is required when the system needs to operate in close proximity, e.g., picking an object and inspection of the ceiling. A numerical investigation for variable aerodynamic effect while the system flies close to the ceiling is given in \cite{robinson2016ceiling}. In a standard control allocation matrix, it is assumed that the change in rotor velocity is proportional to the generated thrust. In order to compensate for the ceiling effect, the real-time data is used in \cite{powers2013influence} for rotor velocity-thrust equations instead of a standard control allocation. This resulted in a better performance as compared to the conventional case. In summary, due to the significant changes in aerodynamic characteristics with the ceiling effect, there has been not enough successful flight demonstration in close proximity mode with the exception of our previous results \cite{kocer2018centralized}, which was based on the nominal conditions.

The available approaches can handle the control of the flying robot when it does not engage with an interaction. However, the challenges associated with the aerodynamic interaction require the system to be more responsive, adaptive and resilient \cite{svacha2017improving,bangura2017thrust,tomic2018simultaneous,yuksel2019aerial}. This operation also brings system and environment based constraints including the level of the interaction. The available approaches that consider the constraints leverage individual multi-models for generic interaction problems which bring additional complexity \cite{alexis2016aerial}. Moreover, nominal optimization-based approaches are considered in the UAV control for the interaction tasks, wherein the system lacks the ability to take external forces, changing parameters and unmodeled dynamics into account \cite{garimella2015towards,lunni_2017,seo2017aerial}.

\section{Problem Formulation}\label{sec_prob_form}

\subsection{Modeling}

A snapshot of the aerial robot in close proximity is presented in Fig. \ref{fig:exp}. In order to define the position of the aerial robot with respect to the world frame, the following transformation can be used
\begin{equation}
\label{eqn:kinematic}
{}^W\pmb{\dot{x}} = R_{WB} {}^B\pmb{\dot{x}},
\end{equation}
where ${}^W\pmb{\dot{x}}$ and ${}^B\pmb{\dot{x}}$ are the translational states in the world frame and the body frame, respectively. $R_{WB}$ is the rotation matrix from the body to the world frame. The second order nonlinear translational dynamics of the aerial robot including external forces is expressed as
\begin{equation}
\label{eqn:dynamics}
\pmb{M}{}^B\ddot{\pmb{x}} - \pmb{\omega}\times{}^B\dot{\pmb{x}} + \pmb{G} = \pmb{F} + \pmb{F}_{\rm ext},        
\end{equation}
where
$\pmb{M}= m \cdot \pmb{I}_3$
is the diagonal mass matrix and $\pmb{\omega}$ is the vector of angular rates. $\pmb{G}$ is gravitational force acting on system in $z$ direction \cite{lee2010geometric}. The input vector is the force generated by the blades on quadrotor in the body frame, $\pmb{F}=[0,0,F_z]^{\rm T}$. Furthermore,  $\pmb{F}_{\rm ext}\in\mathbb{R}^3$ is the external force vector acting on the system.
\subsection{Augmented Formulation}

Consider a nonlinear system, where the system is represented by the sum of a nominal and additive models
\begin{subequations}
\begin{eqnarray}
\dot{\mathrm{x}}(t) &=& f_{n} + f_{a}, \\
f_{n} &=& f\big(\mathrm{x}(t),\mathrm{u}(t)\big), \\
f_{a} &=& \pmb{F}_{\rm ext}.
\label{gen_eqs_dis}
\end{eqnarray}
\end{subequations}
In this representation, the external forces, $\pmb{F}_{\rm ext}=[{F_{\rm ext}}_x, {F_{\rm ext}}_y, {F_{\rm ext}}_z]^{\rm T}$, represent unmodeled dynamics, disturbances, changing parameters as well as the external forces arising during the interaction phase. In this context, we can consider the following problem:

\begin{problem}
	In order to fly in close proximities, how can the system identify external forces precisely? If the system can accurately explore the external forces, how can this data be used within the controller?
\end{problem}

The state vector in the nominal case can be given in terms of the translational positions $(x,y,z)$ and the velocities ($u,v,w$). However, to address the defined problem, we have augmented the nominal case as
%
%
\begin{eqnarray}
\mathrm{x} = [x, y, z, u, v, w, {F_{\rm ext}}_x, {F_{\rm ext}}_y, {F_{\rm ext}}_z]^{\rm T},
\end{eqnarray}
where $\mathrm{x}_k \in \mathbb{X}$. The state constraint set $\mathbb{X}$ is closed, compact and includes the origin. In the augmented model representation, the external force vector is assumed to be a constant disturbance in the form of $\dot{\pmb{F}}_{ext} = 0$. It is also assumed that the origin is included in the feasible set. Therefore, differentially flat states can be driven by the following input vector
%
%
\begin{eqnarray}
\mathrm{u} = [F_z,\phi,\theta,\psi]^{\rm T},
\end{eqnarray}
where $\mathrm{u}_k \in \mathbb{U}$. The control constraint set $\mathbb{U}$ has the same properties with the $\mathbb{X}$. The control vector includes angular positions ($\phi,\theta,\psi$). Therefore, the estimation problem can be formulated in discrete time as 
\begin{subequations}
\begin{eqnarray}\label{eq:model_discrete}
\mathrm{x}_{k+1} &=& f\big(\mathrm{x}_k,\mathrm{u}_k\big) + \mathpzc{w}_k, \\
\mathrm{y}_{k} &=& h\big(\mathrm{x}_k,\mathrm{u}_k\big) + \nu_k,
\end{eqnarray}
\end{subequations}
where the subscript $k$ is the sample taken at time $t_k$, where $\forall k \geqslant 0 $. The function $f(\cdot)$ is composed of discretized versions of \eqref{eqn:kinematic} and \eqref{eqn:dynamics}. In order to obtain \eqref{eq:model_discrete}, a direct multiple shooting method is utilized based on \cite{quirynen2015autogenerating}. For this operation,  Gauss-Legendre integrator of order 4 is preferred with 2 steps per shooting interval and the grid size is chosen 10 ms. Moreover, the physical system parameters are adopted from \cite{kocer_tool}. The  process noise is indicated by $\mathpzc{w}_k \in \mathbb{R}^{N_{\rm x} \times 1}$, where its covariance can be formulated by $\mathbb{E}(\mathpzc{w}\mathpzc{w}^{\rm T})= Q_{\hat{\mathrm{x}}} \in \mathbb{R}^{N_{\rm x} \times N_{\rm x}}$. In order to identify the external forces online, the following measurement function is used 
\begin{equation}
h(\cdot) = \left[x,y,z,u,v,w,F_z,\phi,\theta,\psi\right]^{\rm T}.
\end{equation}
The measurement noise is represented by $\nu_k \in \mathbb{R}^{N_{\rm y} \times 1}$, where its covariance can be formulated by $\mathbb{E}(\nu \nu^{\rm T})= R_{\hat{\mathrm{x}}} \in \mathbb{R}^{N_{\rm y} \times N_{\rm y}}$.
\begin{assumption}
 The noise vectors ($\mathpzc{w}_k$ and $\nu_k$) are independent and normally-distributed random variables.
\end{assumption}

\section{Force Estimation}\label{sec_force_est}

Consider a constrained state estimation problem in the form of a squared norm using the data collected until the jth time step: 
\begin{subequations}\label{eq:inf_est}	
\begin{align}
\min_{\mathrm{x}_k,\mathpzc{w}_k} 
& \sum_{k=0}^{j}\Vert \nu_k \Vert_{V}^2  + \sum_{k=0}^{j-1} \Vert \mathpzc{w}_k \Vert_{W}^2 + \Vert \mathrm{x}_0 - \mathrm{\hat{x}}_0 \Vert_{P_L}^2 \\ 
\text{s.t.} \quad
&\mathrm{x}_{k+1} = f(\mathrm{x}_k,\mathrm{u}_k) + \mathpzc{w}_k, \quad \forall k \in [0,\ldots,j-1]\label{subeq_model} \\
&\mathrm{y}_{k} = h\big(\mathrm{x}_k,\mathrm{u}_k\big) + \nu_k,  \quad \quad \forall k \in [0,\ldots,j]\label{subeq_meas} \\
& \mathrm{x}_{\rm min} \leqslant  \mathrm{x}_k \leqslant \mathrm{x}_{\rm max}\label{subeq_const}
\end{align}
\end{subequations}
where $P_L$ is a positive definite weight matrix to find a balance between initial guess $\mathrm{\hat{x}}_0$ and the initial state $\mathrm{x}_0$. The other positive definite matrices are the inverse of covariance matrices, where $V = Q_{\hat{\mathrm{x}}}^{-1/2}$ and $W = R_{\hat{\mathrm{x}}}^{-1/2}$. Unfortunately, in this generic formulation, the problem may become intractable when the data size increases within time. In order to avoid the curse of dimensionality problem, we can impose a moving window by limiting the number of last measurements. In this context, the estimation window size $N$ is considered, where $L = j-N+1$. The problem in \eqref{eq:inf_est} can be reformulated as follows
\begin{table}[b!]
	\centering
	\small
	\tabcolsep=0.1cm
	\caption {Specification of NMHE.} \label{tab:mhespecs}
	\footnotesize
	\begin{tabular*}{\textwidth}{p{0.48\textwidth}p{0.475\textwidth}}
		\toprule
		\textbf{Parameter}     & \textbf{Value} \\ 
		\rowcolor[rgb]{ .867,  .922,  .969} Time step $\Delta t$ &  $0.01$ (s)  \\ 
		Estimation window $N$ &  $40$  \\ 
		\rowcolor[rgb]{ .867,  .922,  .969}  Measurement noise weight $V$ &  $10^{-4}\cdot{\rm bdiag}(\{25\cdot \pmb{I}_6,\; 0.01,\; 25\cdot \pmb{I}_3\}) $ \\
		Process noise weight $W$ & ${\rm bdiag}(\{1/30\cdot \pmb{I}_6,\; 1/8\cdot \pmb{I}_3\}) $   \\ 
	\rowcolor[rgb]{ .867,  .922,  .969} Arrival cost weight $P_L$ &  ${\rm bdiag}(\{0.01\cdot \pmb{I}_6,\; 0.001\cdot \pmb{I}_3\}) $ \\ 
		Constraints &  $-6 \leqslant {F_z}_{ext} \leqslant 2$  (N) \\ 
		\bottomrule	
	\end{tabular*}
\end{table}
\begin{subequations}\label{eq:fin_est}		
\begin{align}
\min_{\mathrm{x}_k,\mathpzc{w}_k} 
& \sum_{k=L}^{j}\Vert \mathrm{y}_{k} - h\big(\mathrm{x}_k,\mathrm{u}_k\big) \Vert_{V}^2  + \sum_{k=L}^{j-1} \Vert \mathpzc{w}_k \Vert_{W}^2 + \Vert \mathrm{\hat{x}}_L - \mathrm{\bar{x}}_L \Vert_{P_L}^2 \\ 
\text{s.t.}  \quad
&\mathrm{x}_{k+1} = f(\mathrm{x}_k,\mathrm{u}_k) + \mathpzc{w}_k,   \quad \forall k \in [0,\ldots,j-1] \\
&\mathrm{y}_{k} = h\big(\mathrm{x}_k,\mathrm{u}_k\big) + \nu_k,  \quad \quad \forall k \in [0,\ldots,j] \\
& \mathrm{x}_{\rm min} \leqslant  \mathrm{x}_k \leqslant \mathrm{x}_{\rm max}
\end{align}
\end{subequations}	
%
%
where $\mathrm{\bar{x}}$ is the estimation value given by the arrival cost to approximate the past values until the first sample of the estimation window. In terms of the estimation problem, this term is similar to the initial guess $\mathrm{\hat{x}}_0$ in \eqref{eq:inf_est} since it is the first term of the estimation window.  Similarly, $\mathrm{\hat{x}}$ is the estimation value given by the moving horizon estimation. 
The arrival cost can be defined by	Eq. \eqref{eq:arr_est}
\begin{subequations}\label{eq:arr_est}	
\begin{align}
\argmin_{\mathrm{x}_k,\mathpzc{w}_k} 
& \sum_{k=-\infty }^{L}\Vert \mathrm{y}_{k} - h\big(\mathrm{x}_k,\mathrm{u}_k\big) \Vert_{V}^2  + \sum_{k=-\infty}^{L-1} \Vert \mathpzc{w}_k \Vert_{W}^2 \\ 
\text{s.t.}  \quad
&\mathrm{x}_{k+1} = f(\mathrm{x}_k,\mathrm{u}_k) + \mathpzc{w}_k, \forall k \in [0,\ldots,j-1]\\
&\mathrm{y}_{k} = h\big(\mathrm{x}_k,\mathrm{u}_k\big) + \nu_k,  \quad \forall k \in [0,\ldots,j]
\end{align}
\end{subequations}	
where it can be solved with linearity assumptions, e.g., a Kalman filter. The solution of \eqref{eq:arr_est} is adopted from \cite{kuhl_2011}. The output of \eqref{eq:arr_est} will be $\mathrm{\bar{x}}_{L+1}$ and $P_{L+1}$ for the next iteration in \eqref{eq:fin_est}. The specified parameters for the estimation problem are summarized in Table \ref{tab:mhespecs}.

%
%

%
\begin{assumption}
The state function $f(\cdot)$, and the associated costs are continuous and differentiable.
\end{assumption}
%


\section{Controller Design}\label{sec_controller}
\begin{figure}[t]
	\centering
	\includegraphics[width=\columnwidth]{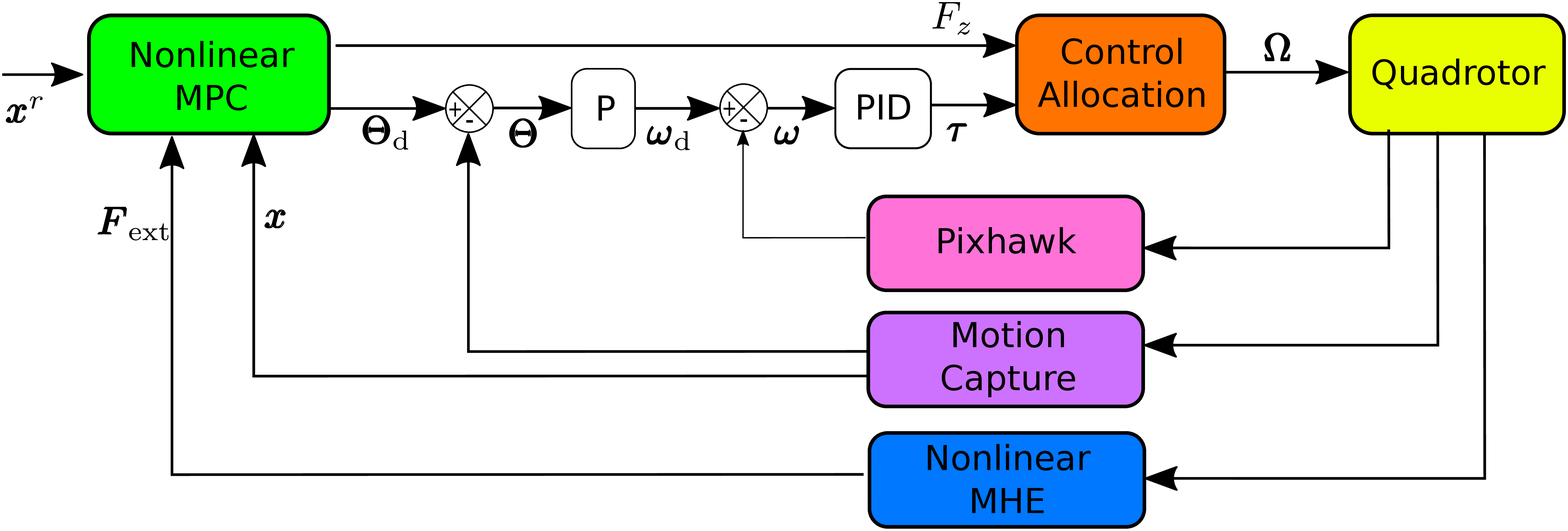}
	\caption{The closed loop scheme: a force estimation-based NMPC approach. The output of the NMPC for the attitude angles are given in $\pmb{\Theta}_{\text{d}}$. After PID controller loops, the torques $\pmb{{\tau}}$ and vertical force values $F_z$ are fed into the control allocation matrix. The rotor velocities are given in $\pmb{{\Omega}}$.} 
	\label{fig:controller_diagram}
\end{figure}

In this work, two staged feedback controller is implemented. In the first stage; the NMPC, which generates the force in $z$ direction and attitude reference for the feedback controller, is used. In the second stage, a cascaded P and PID controller is used to generate desired momentum to be applied by rotors. Finally, the rotor speeds are calculated by control allocation matrix. The proposed approach is illustrated in Fig. \ref{fig:controller_diagram}.

\begin{table}[b!]
	\rowcolors{2}{}{Wheat1}
	\centering
	\small
	\tabcolsep=0.1cm
	\caption {Specification of NMPC.} \label{tab:mpcspecs}
	\footnotesize
	\begin{tabular*}{\textwidth}{p{0.48\textwidth}p{0.475\textwidth}}
		\toprule
		\textbf{Parameter}     & \textbf{Value} \\ 
		\rowcolor[rgb]{ .867,  .922,  .969}  Time step $\Delta t$ &  $0.01$ (s)   \\ 
		Prediction and control horizon $N$ &  $40$  \\ 
		\rowcolor[rgb]{ .867,  .922,  .969}  Stage cost weight $Q$ &  $diag(30, 30, 10, 1, 1, 2.5)$  \\ 
		Input weight $R$ & $diag(30, 30, 80, 4\times 10^{-2})$  \\ 
		\rowcolor[rgb]{ .867,  .922,  .969}  Terminal cost weight $S$ &  $diag(60, 60, 20, 2, 2, 5)$  \\ 
		Constraints &  $0.5 \leqslant F_z \leqslant 1.5$ (mg)  \\ 
		\bottomrule	
	\end{tabular*}
\end{table}
Consider an optimization-based control problem in the form of a squared norm:
\begin{subequations}	
\begin{align}\label{eq:inf_cont}
\min_{\mathrm{x}_k,\mathpzc{u}_k} 
& \sum_{k=j}^{\infty}\Vert e_{\mathrm{x}} \Vert_{Q}^2  + \sum_{k=j}^{\infty} \Vert e_{\mathrm{u}} \Vert_{R}^2 \\ 
\text{s.t.} \quad
&\mathrm{\hat{x}}_k=\mathrm{{x}}_k \label{subeq_estconst} \\
&\mathrm{x}_{k+1} = f(\mathrm{x}_k,\mathrm{u}_k) \label{subeq_modelmpc}  \\ 
&\mathrm{x}_k = (\mathrm{x}_0,\mathrm{x}_1, \ldots, \mathrm{x}_{k-1}, \mathrm{x}_k,\dots)  \label{subeq_states}  \\ 
&\mathrm{u}_k = (\mathrm{u}_0,\mathrm{u}_1, \ldots, \mathrm{u}_{k-2}, \mathrm{u}_{k-1}, \dots)  \label{subeq_controls}  \\ 	
& \mathrm{x}_{\rm min} \leqslant  \mathrm{x}_k \leqslant \mathrm{x}_{\rm max} \label{subeq_constmpc} \\ 	
& \mathrm{u}_{\rm min} \leqslant  \mathrm{u}_k \leqslant \mathrm{u}_{\rm max} \label{subeq_constmpc2}
\end{align}
\end{subequations}	
where $e_{\mathrm{x}}=(\mathrm{x}_k^r-\mathrm{x}_k)$ and $e_{\mathrm{u}}=(\mathrm{u}_k^n-\mathrm{u}_k)$ in which $\mathrm{x}_k^r$ is the trajectory reference for the system and $\mathrm{u}_k^n$ is the nominal control signal. The weight matrix $Q$ is positive semidefinite and the weight matrix $R$ is positive definite. These weight matrices might affect the performance of the system. However, for this set of controlled states by the infinite sequences of control actions, the problem may not be applicable due to the potential infinite dimensional optimization problem \cite{eren2017model}. Similar to the NMHE case, the infinite dimensional problem can be defined in a receding horizon manner:
\begin{subequations}	
\begin{align}\label{eq:inf_cont}
\min_{\mathrm{x}_k,\mathpzc{u}_k} 
& \sum_{k=j}^{j+N-1} \Big( \Vert e_{\mathrm{x}} \Vert_{Q}^2  +  \Vert e_{\mathrm{u}} \Vert_{R}^2 \Big) + \sum_{k=j}^{j+N} \Vert e_{\mathrm{x}} \Vert_{S}^2  \\ 
\text{s.t.} \quad 
&\eqref{subeq_estconst},\eqref{subeq_modelmpc},\eqref{subeq_states},\eqref{subeq_controls},\eqref{subeq_constmpc},\eqref{subeq_constmpc2}
\end{align}
\end{subequations}	
In this representation, the contribution of the states and control actions beyond the finite and moving horizon is approximated by the terminal cost. Similar to the stage cost weight, the weight matrix $S$ is also positive semidefinite. The specified parameters for the controller are given in Table \ref{tab:mpcspecs}.


The proposed optimization-based approach is set using real-time iteration scheme in ACADO \cite{houska2011acado} and solved by qpOASES \cite{ferreau2014qpoases}. First, the self-contained C codes are generated by ACADO for the NMHE and NMPC. Afterward, these codes are integrated into the ROS-Kinetic environment. With the evaluations in the simulation environment (Gazebo), the system is tested in OptiTrack motion capture system, which localizes the aerial robot at 240 Hz over the wifi network. 

\begin{figure}[t!]
	\centering
	\includegraphics[width=\columnwidth]{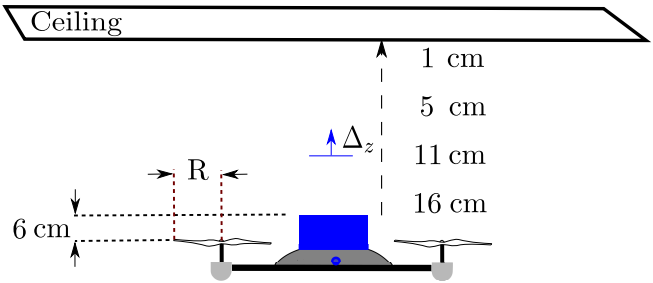}
	\caption{Aerial robot in a close proximity.}
	\label{fig:ceiling_scheme}
\end{figure}

\section{Experimental Results}
\label{sec_experiments}

In order to show the effectiveness of our proposed approach, we performed a set of experiments. By leveraging the optimization-based force estimation and the optimization-based controller, the quadrotor system is tested while approaching the ceiling. 

\subsection{Experimental Setup}

\begin{figure}[t]
	\includegraphics[width=1\textwidth]{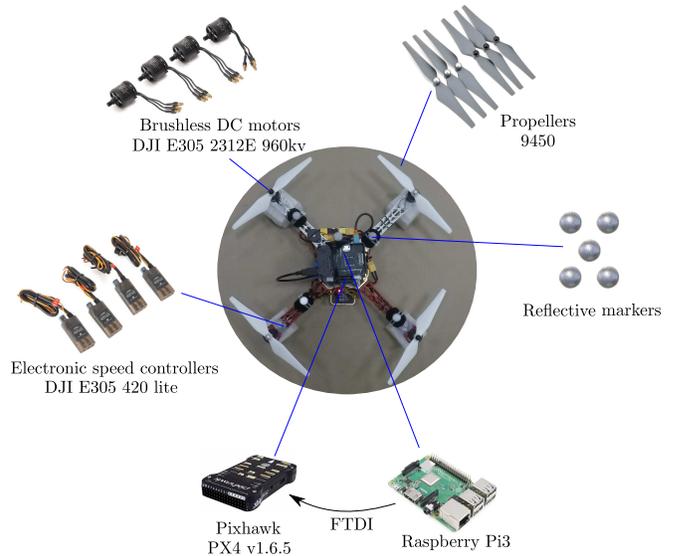}
	\caption{Subcomponents of the aerial robot. }
	\label{fig:robot_components}	
\end{figure}

We used a quadrotor platform for the experiments and the scheme for the test procedure can be seen in Fig. \ref{fig:ceiling_scheme}. It is a small scale quadrotor (DJI F450) and its subcomponents can be seen in Fig. \ref{fig:robot_components}. This experimental setup is equipped with a PX4 FMU and a Raspberry Pi 3 onboard computer unit. While the Raspberry Pi 3 is responsible for the higher-level tasks (commanding generated throttle and angular positions), the PX4 FMU (Firmware v1.6.5) handles the attitude setpoint tracking as well as reaching required vertical force values. For the serial connection between onboard computer and PX4, an FTDI cable is used. In the experiments, the PX4 unit's attitude controller is used. 
\begin{figure}[b]
	\includegraphics[width=1\textwidth]{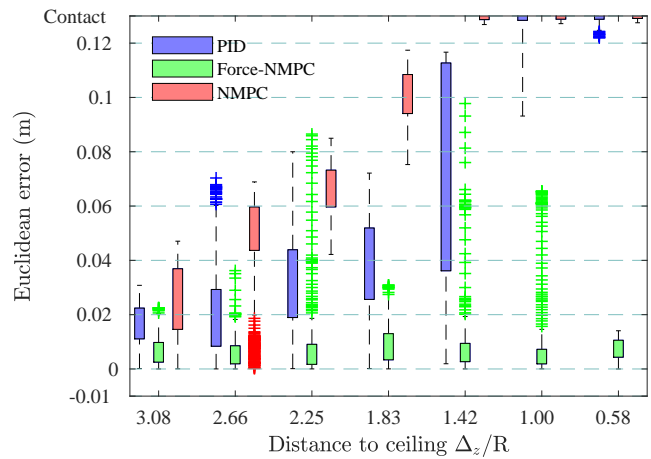}
	\caption{Performance of different controllers in ceiling proximity experiment. x axis shows the desired vertical distance from ceiling in decreasing order, and y axis is the Euclidean error. Bars in different color represent standard deviation of error for each controller. Small error bar for $1.00$ and $0.58$ indicates that robot stick on the ceiling. }
	\label{fig:ceilingstat}	
\end{figure}

\subsection{Close Proximity Flight Performance Comparison}

In order to evaluate the performance of the proposed controller statistically, we have compared its performance with PID and NMPC without force estimation in a scenario, where the robot is commanded to stabilize itself under a ceiling in different proximities. The deviation from the desired distance to ceiling is measured for three controllers and then is plotted in Fig. \ref{fig:ceilingstat}. The results show that the level of interaction starts affecting the performance of the PID and NMPC while approaching the ceiling. They could not bring the robot back to the reference point. The performance of these controllers further deteriorates as the robot approaches the ceiling and eventually they fail to overcome the suction force and the robot stick on the ceiling. On the other hand, the proposed force estimation-based nonlinear MPC approach manages to keep the robot even under 1 cm below the ceiling without experiencing any sticking.

\subsection{Evaluation of the Battery Performance in Close Proximity}

\begin{figure}[t]
	\includegraphics[width=1\columnwidth]{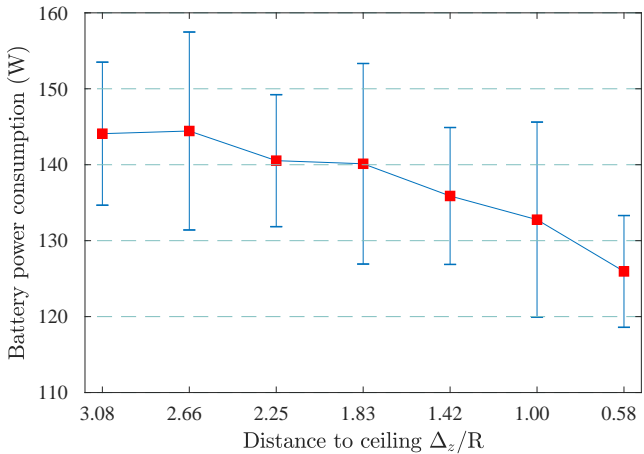}
	\caption{Battery power consumption for different proximity to ceiling. x axis shows proximity to ceiling in decreasing order, and y axis shows average current consumption.}
	\label{fig:batterystat}	
\end{figure}

We investigated battery currents and voltages measured by Pixhawk to evaluate the power efficiency of the flight. In this analysis, we excluded PID and NMPC since they already fail to stay in close proximity interaction zones. The average power consumption is calculated as 
\begin{equation}
    P_{ave} = \frac{1}{T}\int_{t=0}^{t=T}v(t)i(t)dt,
\end{equation}
where $T$ is the duration of measurement for each distance, and $v(t)$ and $i(t)$ are  battery voltage and drawn current at time $t$, respectively. The experimental average power consumption results are summarized in Fig. \ref{fig:batterystat} for different proximities. It is observed that the power consumption of the system decreases up to $12.5\%$ in close proximity flight. Considering the declined power demand of the UAV, we expect that the flight duration can be longer when the system flies below the ceiling. 
%

\subsection{Further Controller Performance Analysis}

\begin{figure*}[htp]
	\subfloat[The system performance on vertical axis.]
	{
		\includegraphics[width=0.333\columnwidth]{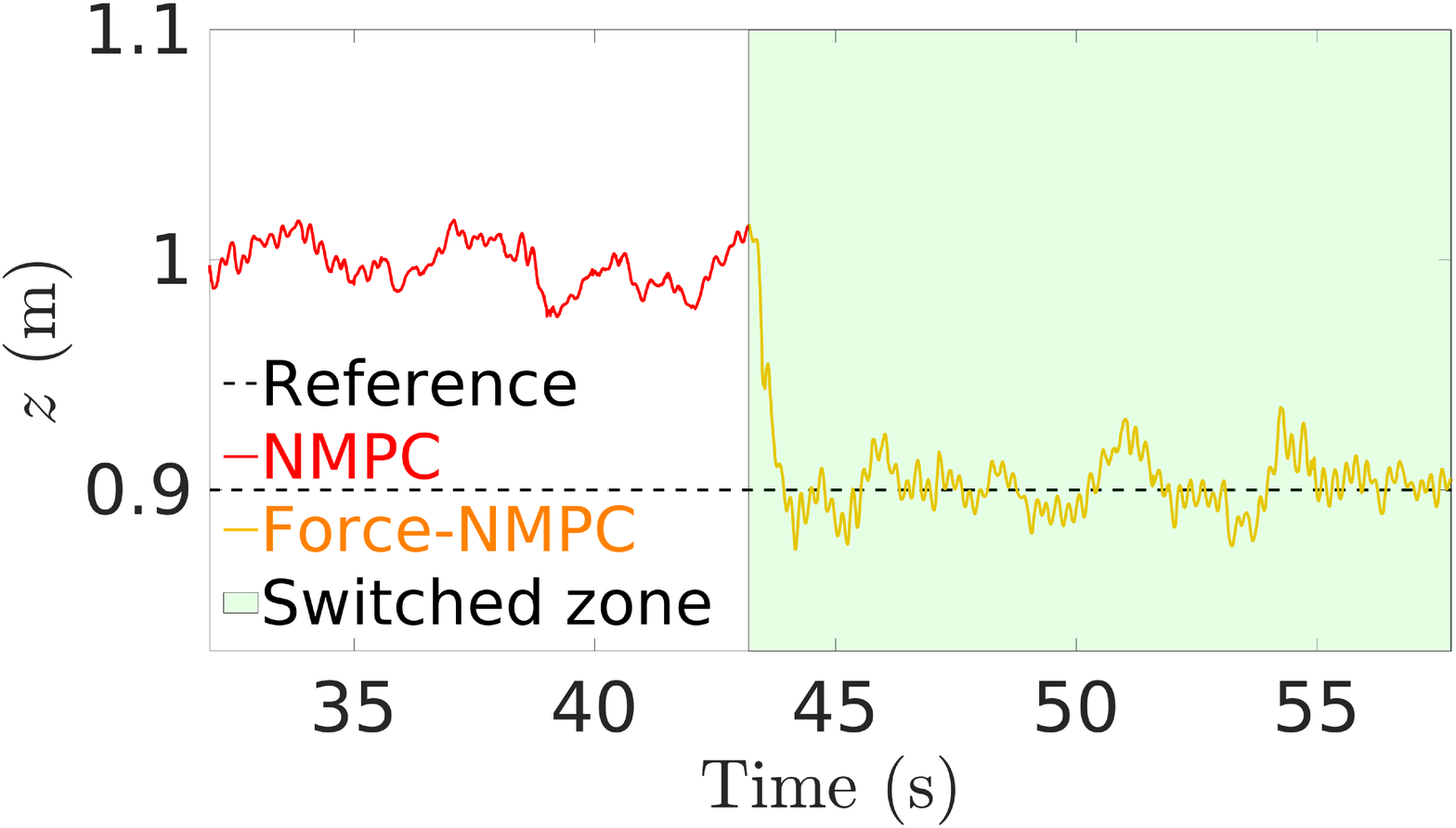}
		\label{fig_z_16}
	} 
	\subfloat[Control effort.]
	{
		\includegraphics[width=0.333\columnwidth]{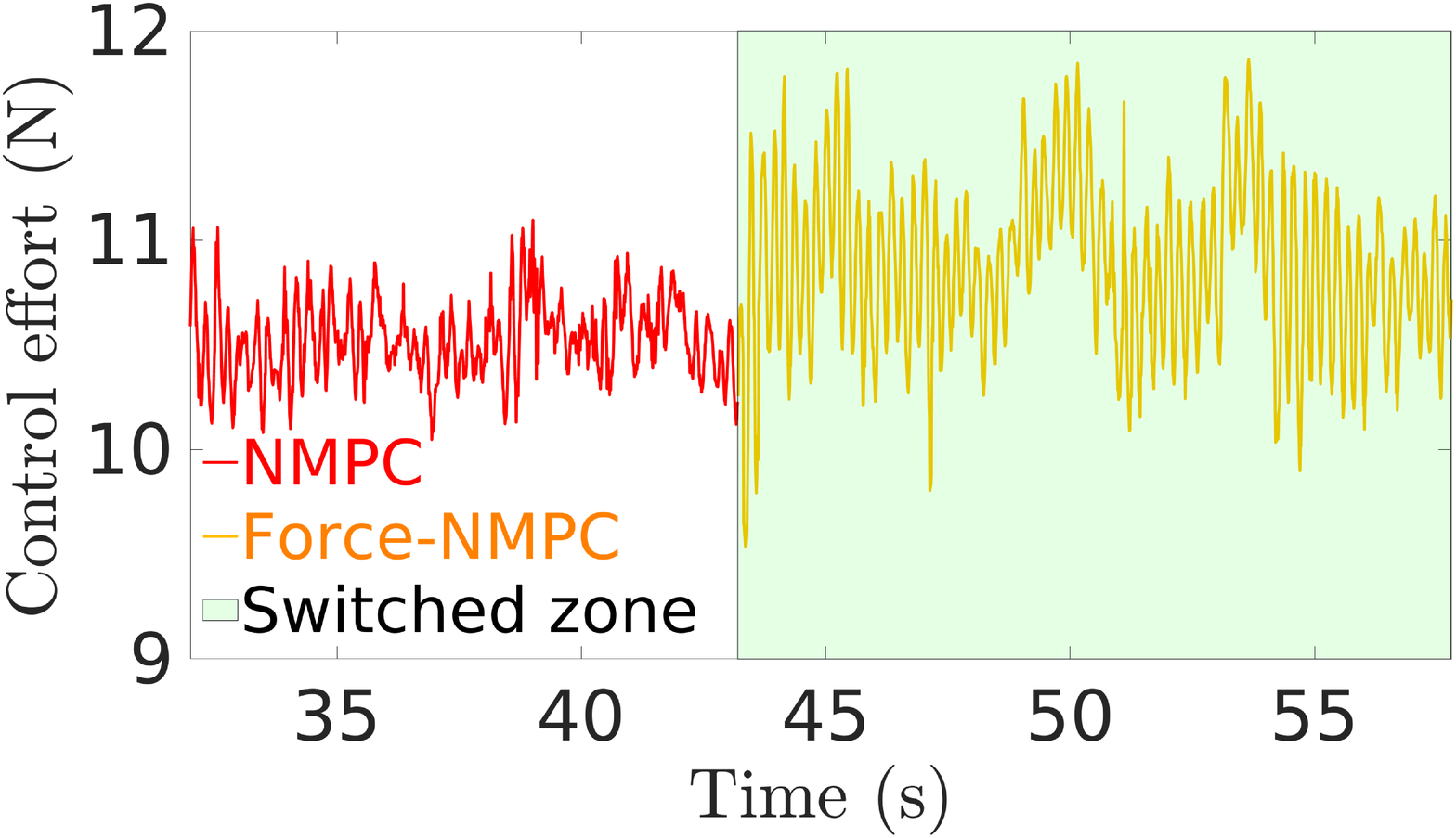}
		\label{fig_c_16}
	} 
	\subfloat[Force estimation.]
	{
		\includegraphics[width=0.333\columnwidth]{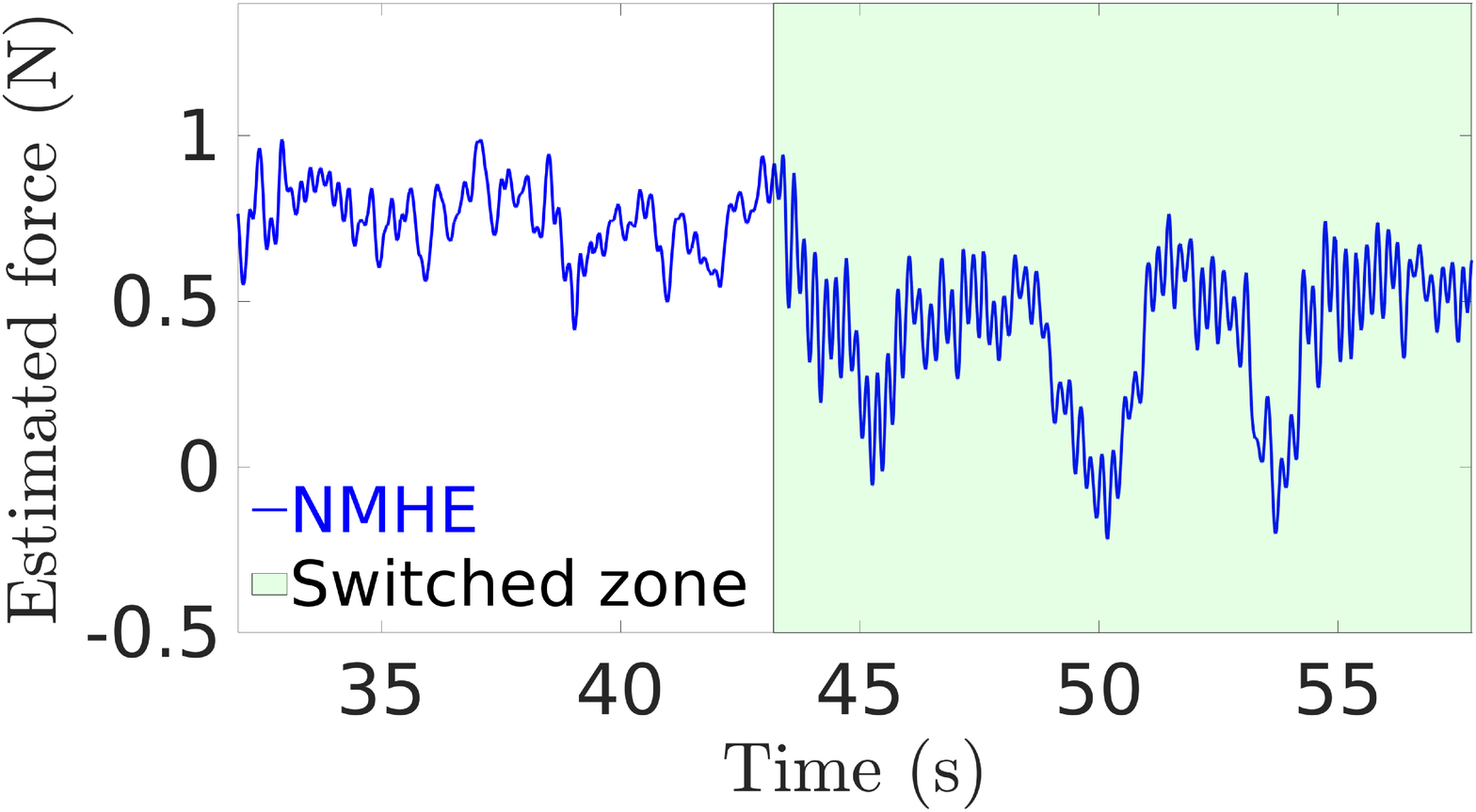}
		\label{fig_f_16}
	}
	\caption{The system performance for $\Delta_z/{\rm R} = 1.83$.}
	\label{ceiling16}	
\end{figure*}

In the flight tests, a random point in the air is set for the aerial robot. After this hovering phase, a reference is generated to go below the ceiling. In order to evaluate system performance, different proximities are defined. For this set of tests, the distance is measured from the top of the Raspberry Pi, as it is illustrated in Fig. \ref{fig:ceiling_scheme}. The distance between the propellers and the Raspberry Pi is 6 cm. In the first experiment, the system needs to stay by keeping its orientation below 16 cm from the ceiling. In Fig. \ref{ceiling16}, the system response on the tracking (Fig. \ref{fig_z_16}), controller action (Fig. \ref{fig_c_16}) and the force estimation (Fig. \ref{fig_f_16}) can be seen. The first instant of the green region indicates the switching mechanism, where the force estimation-based NMPC is activated. In the first part of the figures, the nominal model-based NMPC is used. As it can be seen, the controller can bring the system to the defined reference when the additive model is leveraged. 

\begin{figure*}[htp]
	\subfloat[The system performance on vertical axis.]
	{
		\includegraphics[width=0.333\columnwidth]{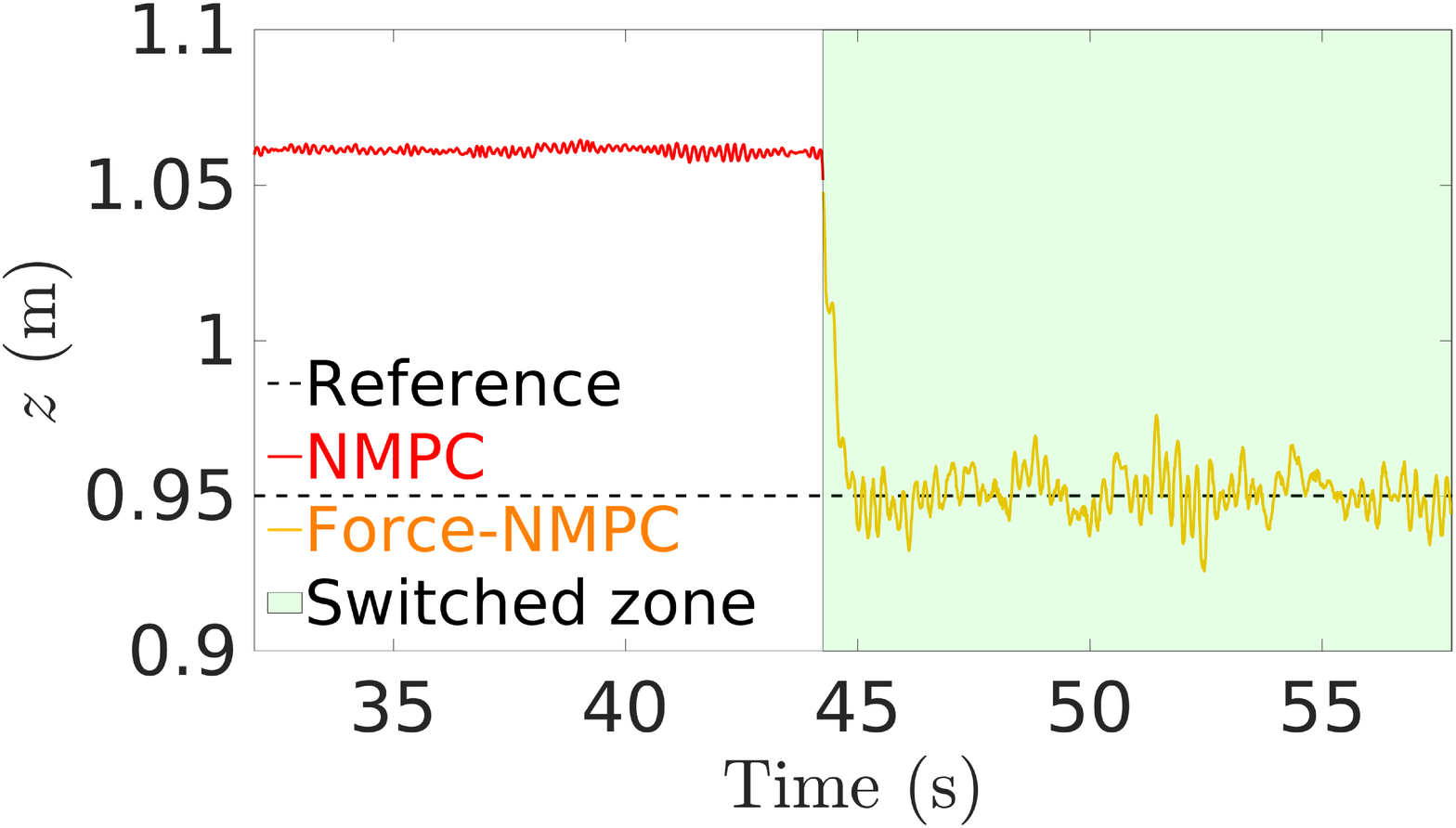}
		\label{fig_z_11}
	} 
	\subfloat[Control effort.]
	{
		\includegraphics[width=0.333\columnwidth]{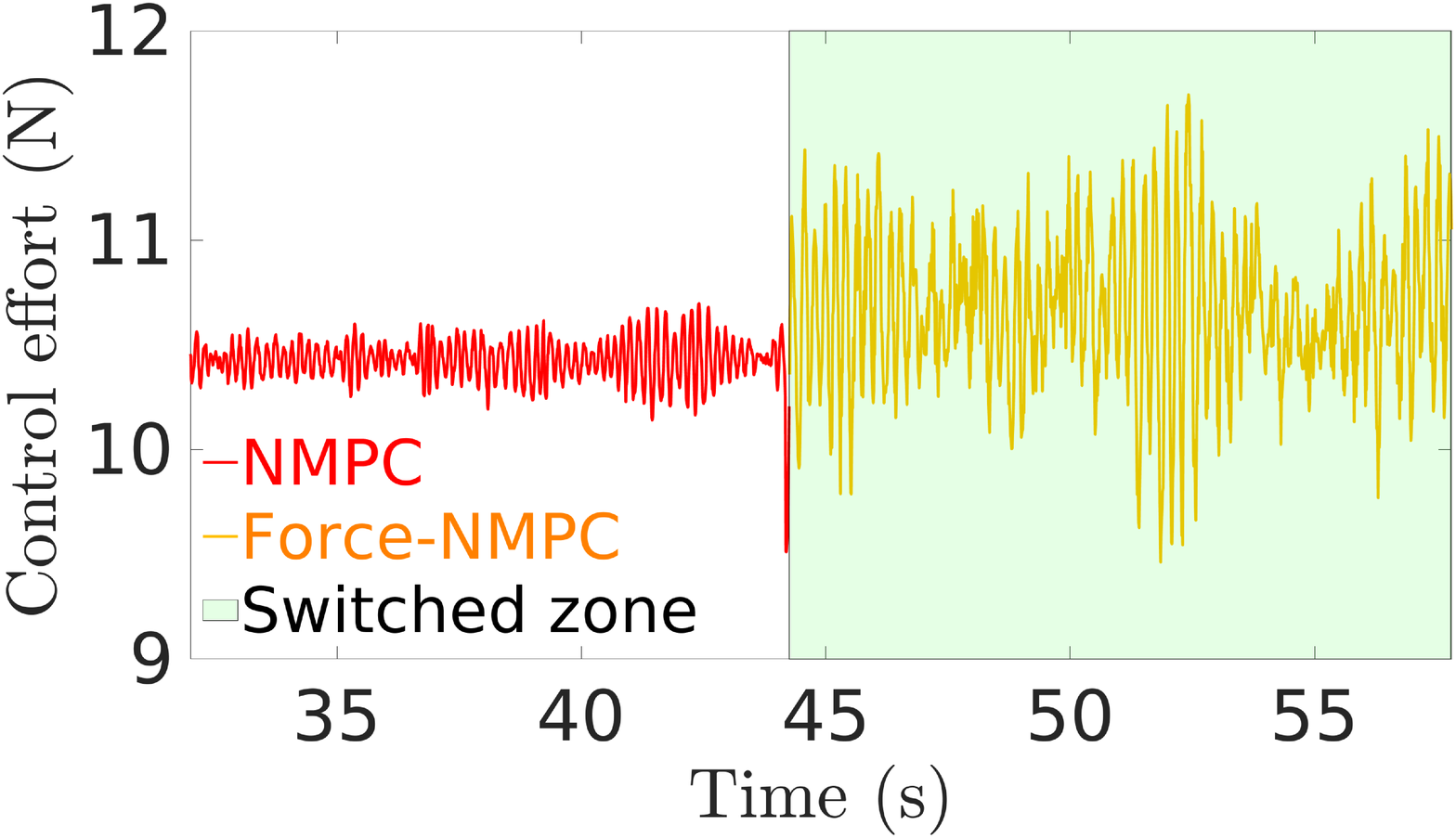}
		\label{fig_c_11}
	} 
	\subfloat[Force estimation.]
	{
		\includegraphics[width=0.333\columnwidth]{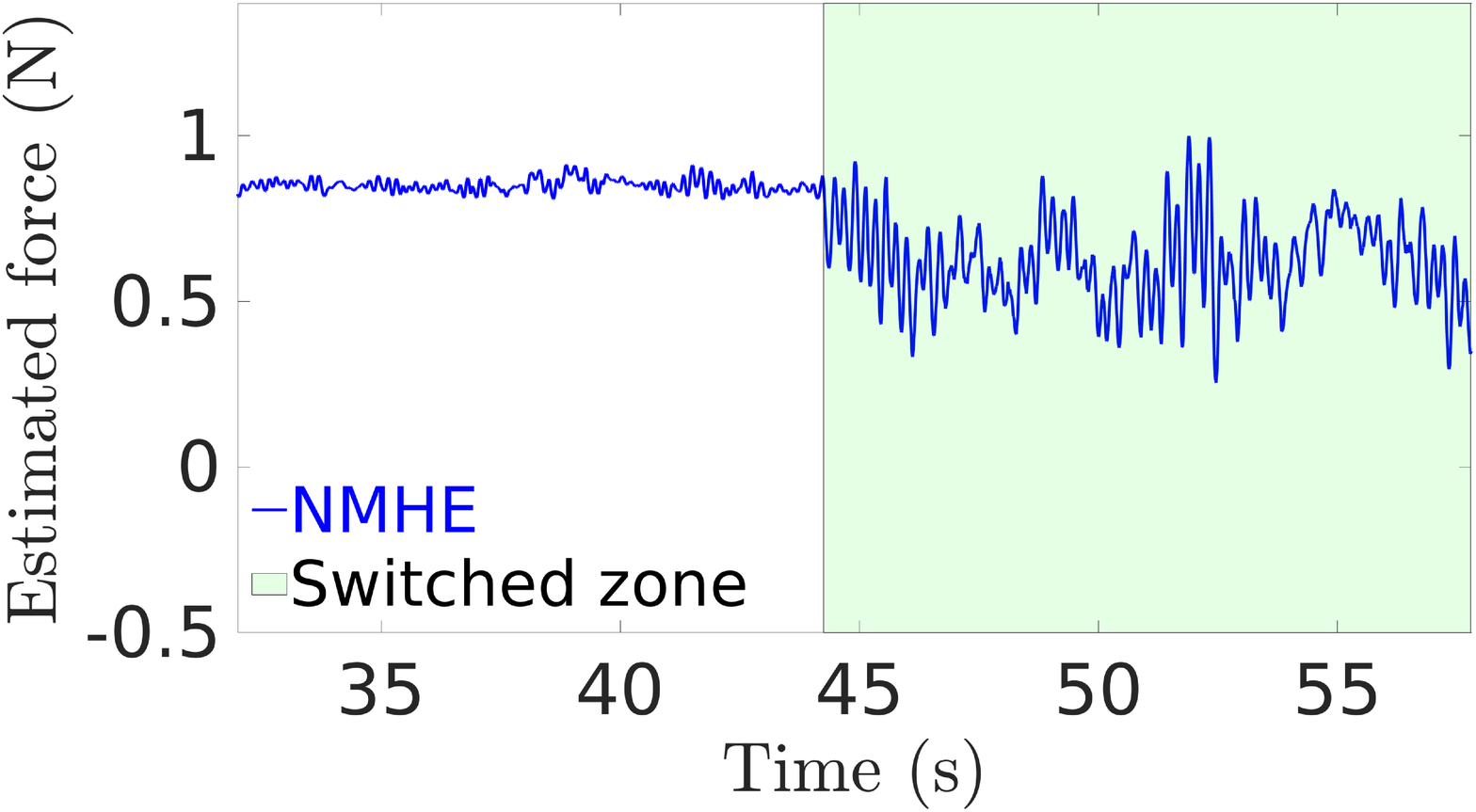}
		\label{fig_f_11}
	}
	\caption{The system performance for $\Delta_z/{\rm R} = 1.42$.}
	\label{ceiling11}	
\end{figure*}

Similar to the first case, the system is also tested below 11 cm from the ceiling. When the switching mechanism is activated, the proposed approach actively suppress the ceiling effect. It is noted that the disturbance on the system becomes more dominant while it flies within 10 cm range as can be seen from Fig. \ref{ceiling11}. 

\begin{figure*}[htp]
	\subfloat[The system performance on vertical axis.]
	{
		\includegraphics[width=0.333\columnwidth]{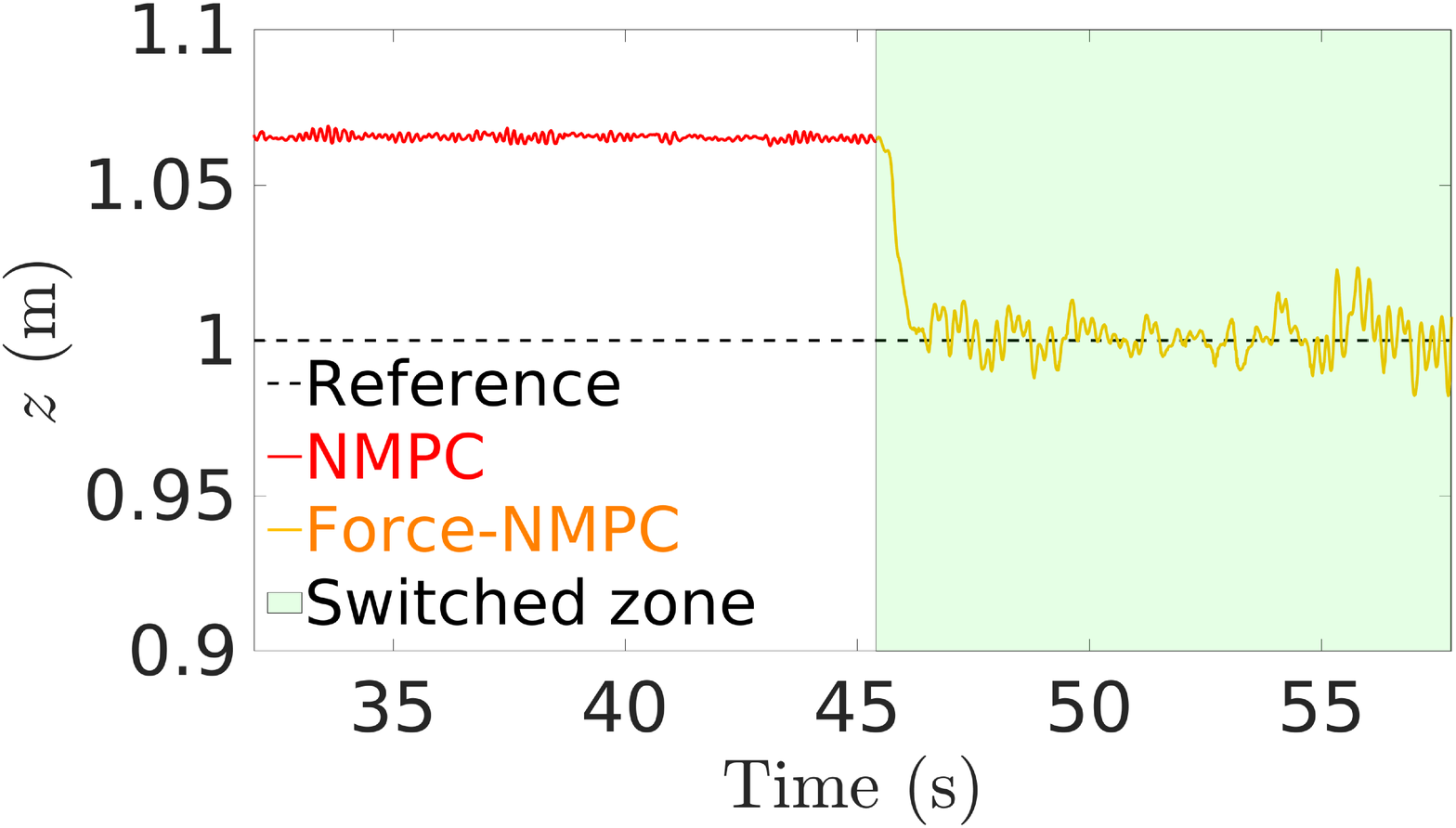}
		\label{fig_z_6}
	} 
	\subfloat[Control effort.]
	{
		\includegraphics[width=0.333\columnwidth]{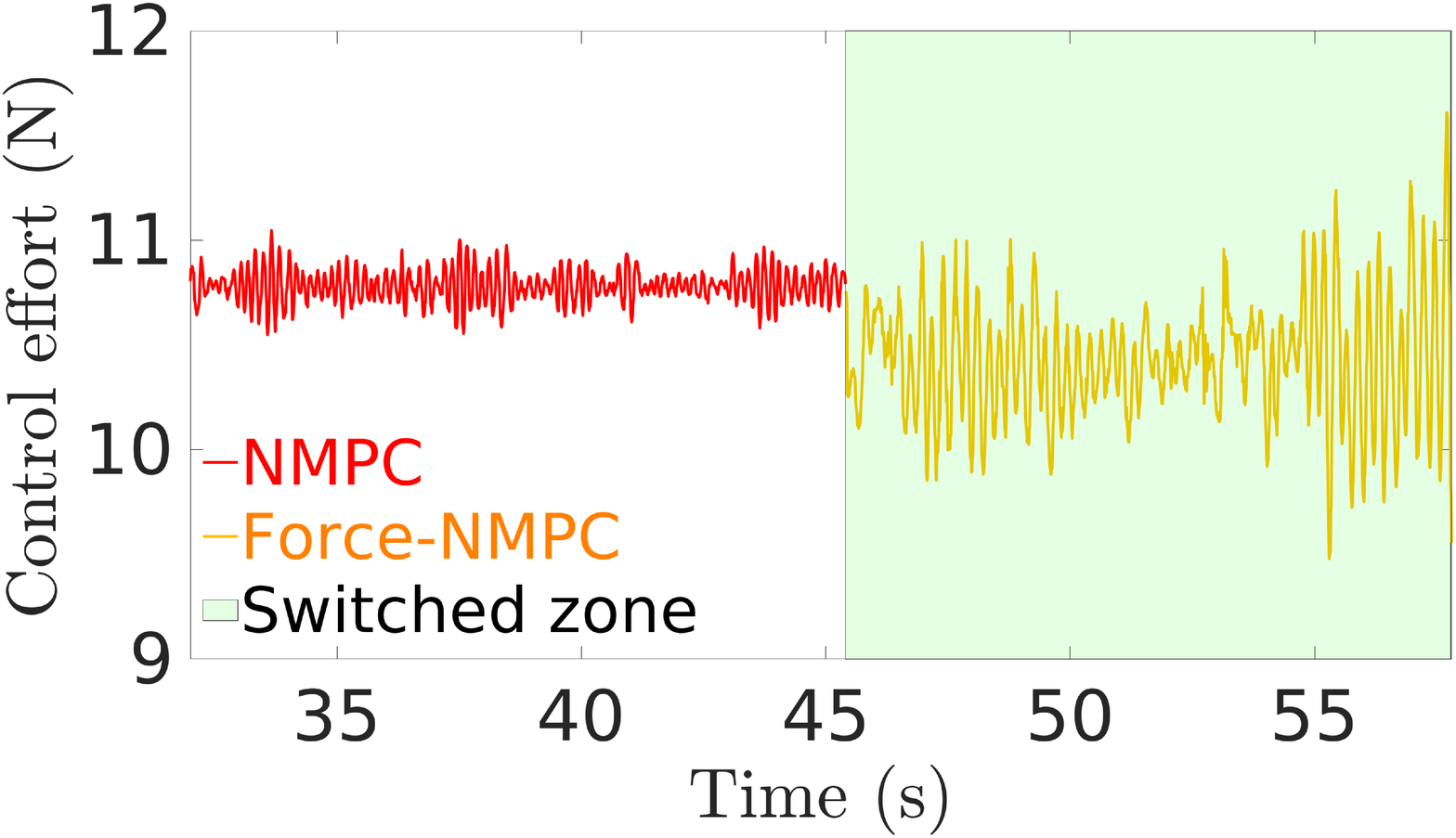}
		\label{fig_c_6}
	} 
	\subfloat[Force estimation.]
	{
		\includegraphics[width=0.333\columnwidth]{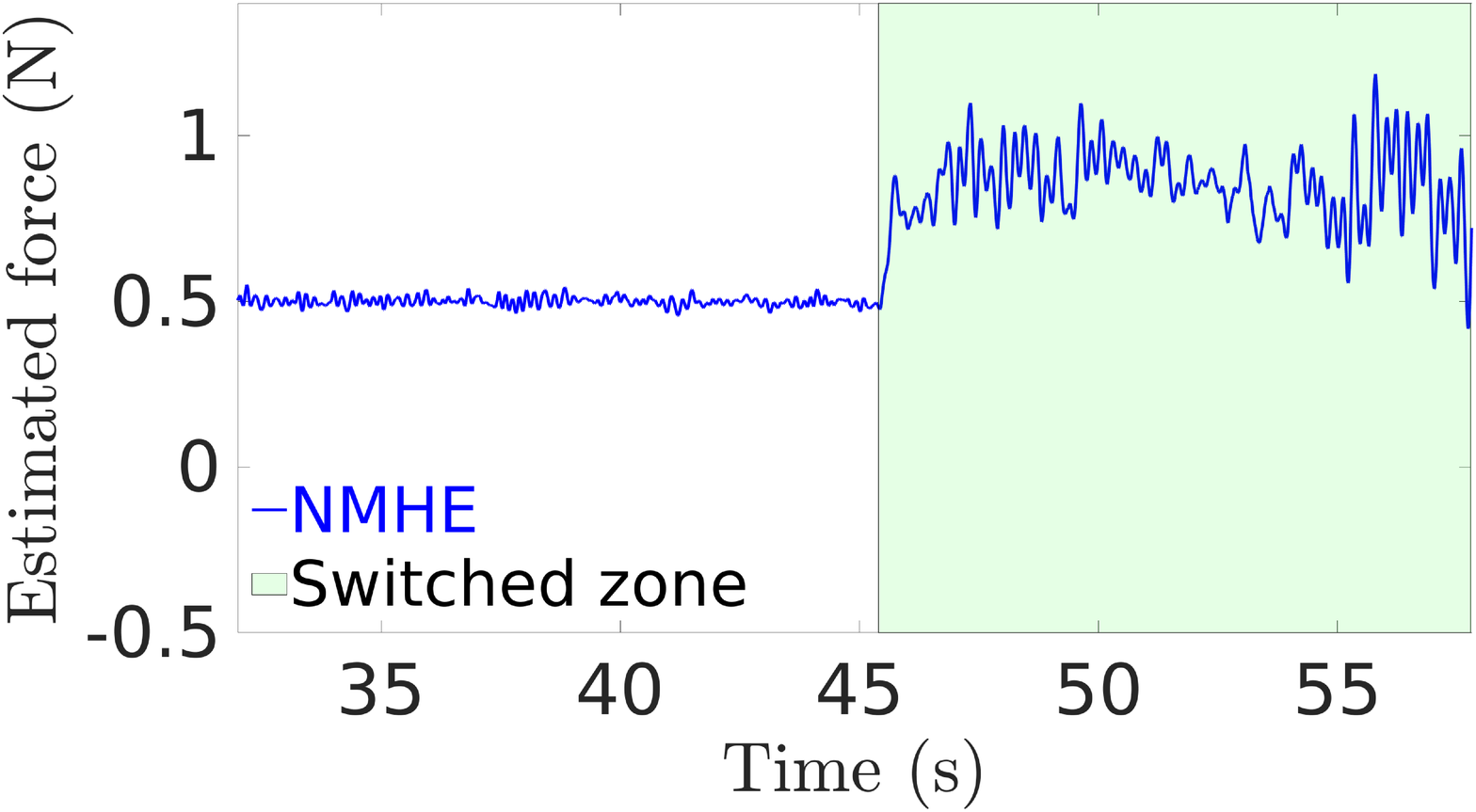}
		\label{fig_f_6}
	}
	\caption{The system performance for $\Delta_z/{\rm R} = 1.00$.}
	\label{ceiling6}	
\end{figure*}

The system performance below 6 cm from the ceiling is given in Fig. \ref{ceiling6}. The active force estimation-based NMPC mitigates the ceiling effect. As compared to Fig. \ref{fig_c_16} and Fig. \ref{fig_c_11}, the controller effort is decreased. Since the ceiling effect increases the rotor wake which results in an increase in the thrust; to stay in close proximities to the ceiling, the system does not need to generate the same thrust when hovering in the free flight case

\begin{figure*}[htp]
	\subfloat[The system performance on vertical axis.]
	{
		\includegraphics[width=0.333\columnwidth]{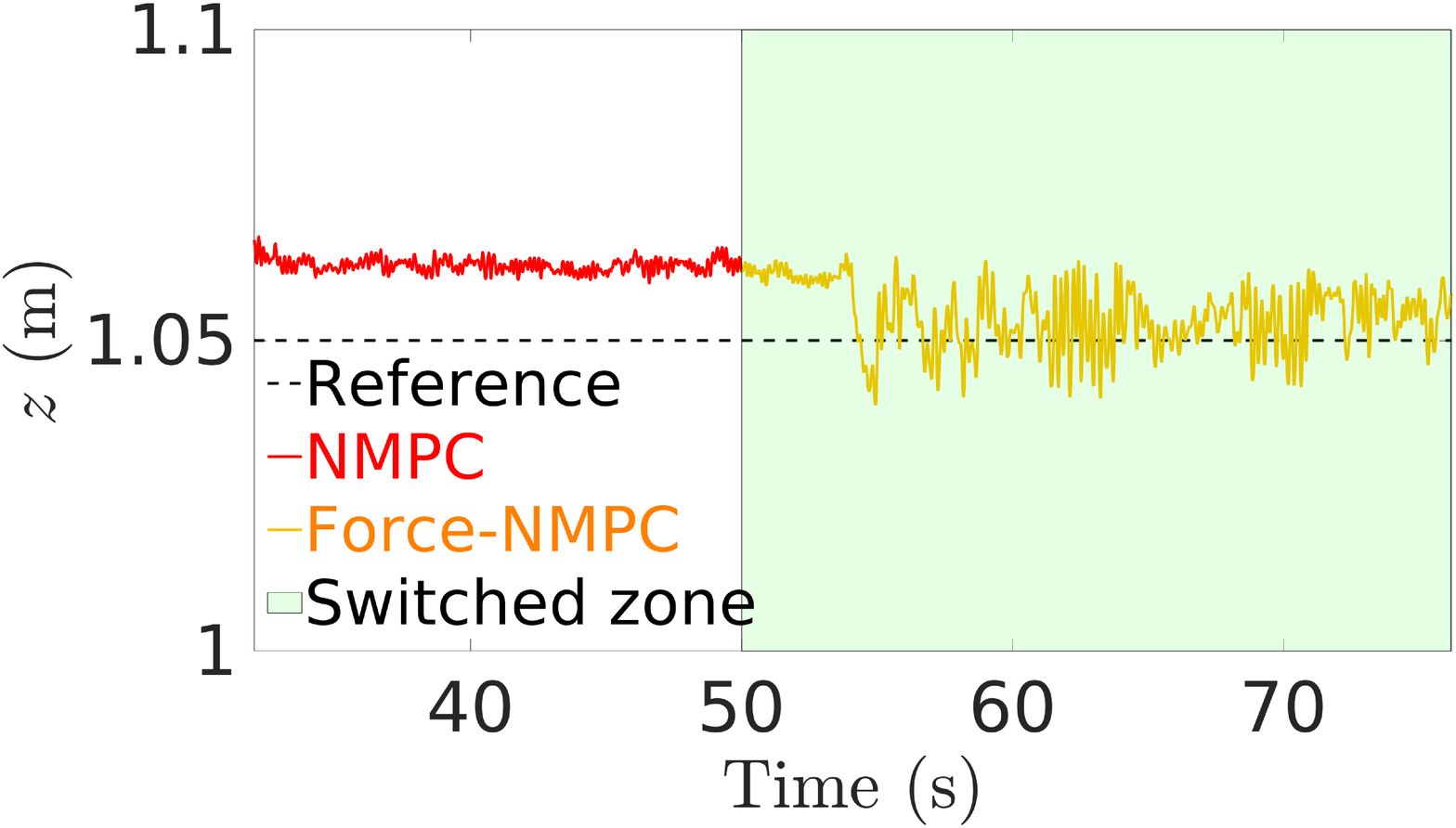}
		\label{fig_z_1}
	} 
	\subfloat[Control effort.]
	{
		\includegraphics[width=0.333\columnwidth]{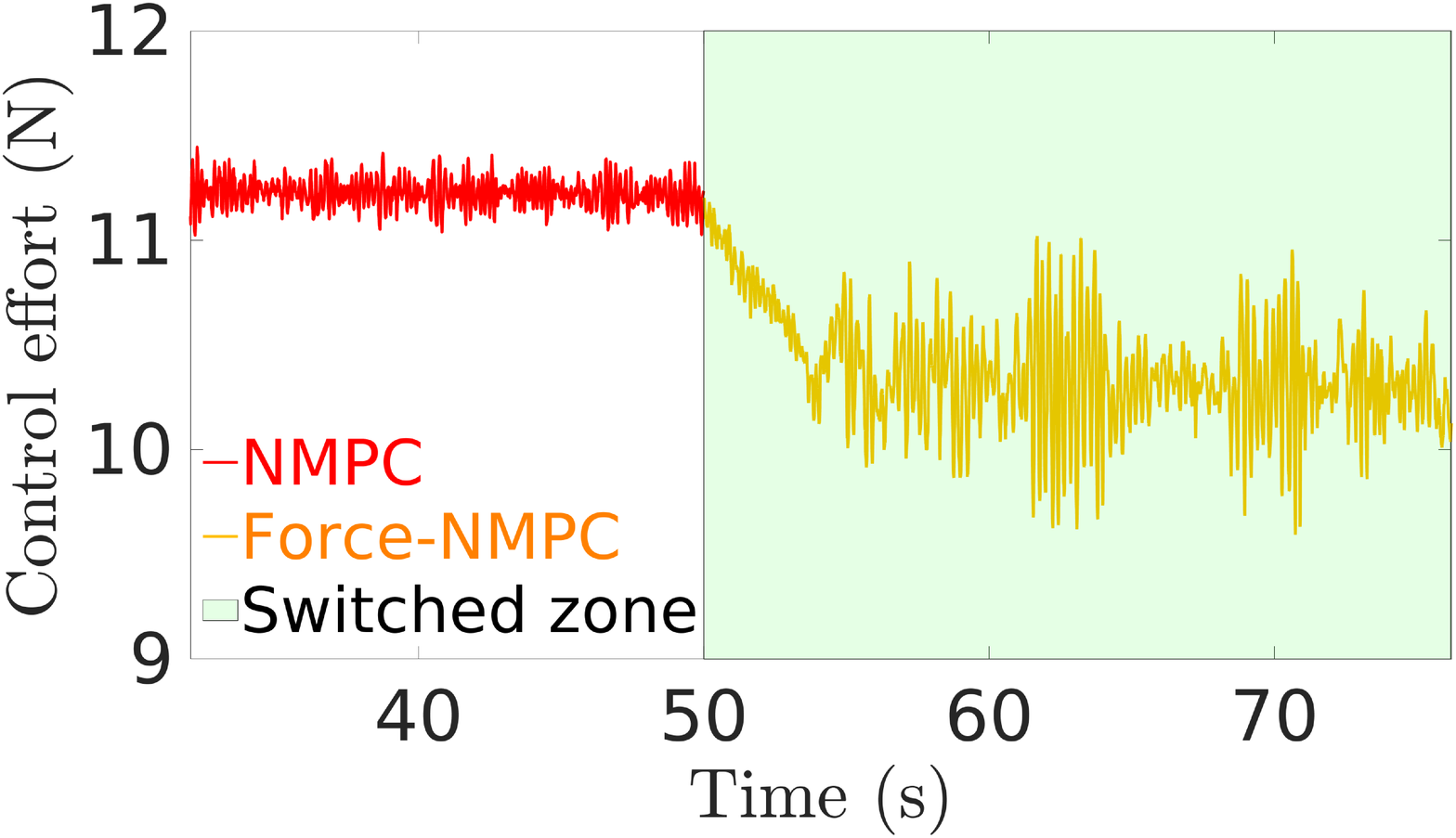}
		\label{fig_c_1}
	} 
	\subfloat[Force estimation.]
	{
		\includegraphics[width=0.333\columnwidth]{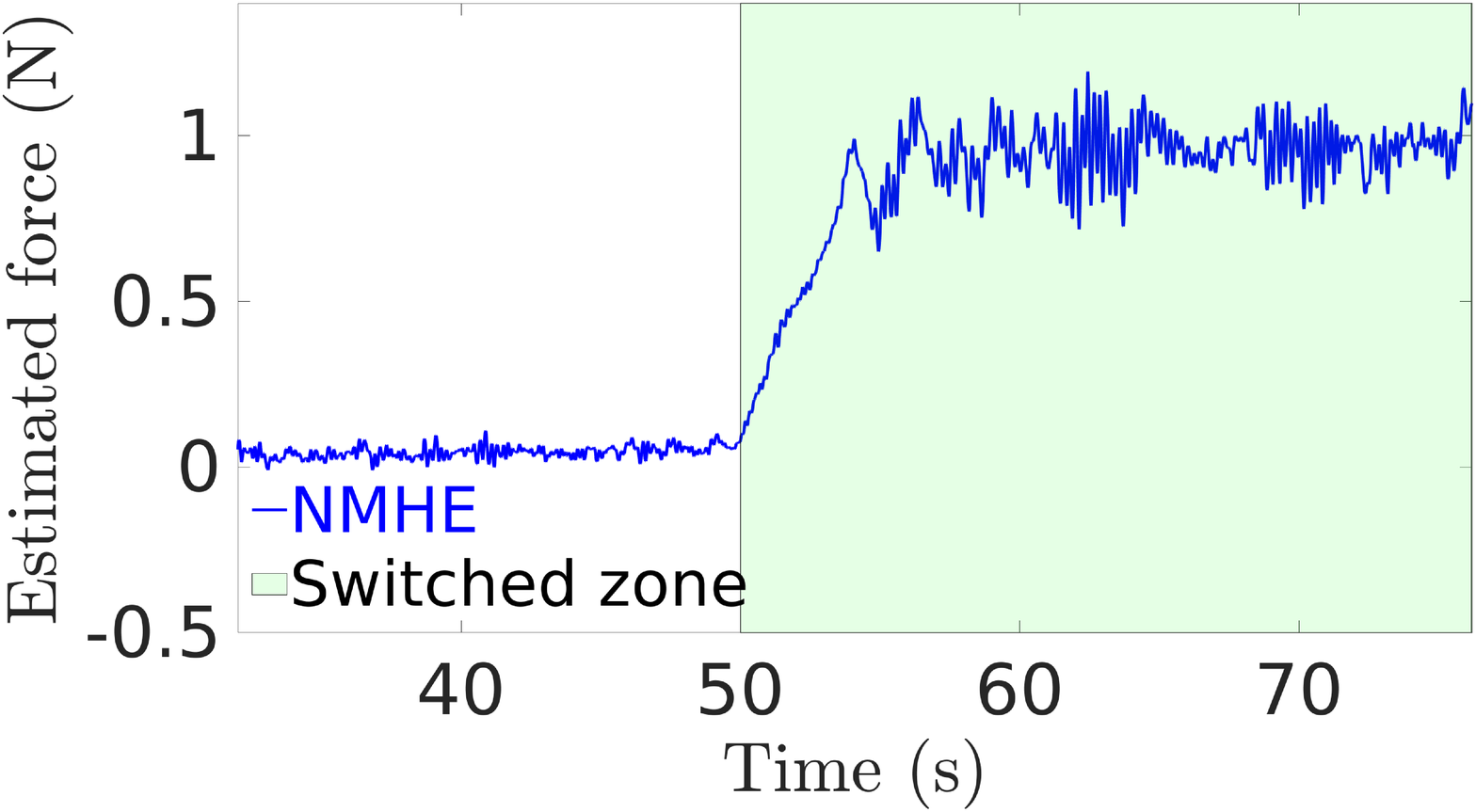}
		\label{fig_f_1}
	}\\
	\subfloat[Battery state.]
	{
		\includegraphics[width=0.333\columnwidth]{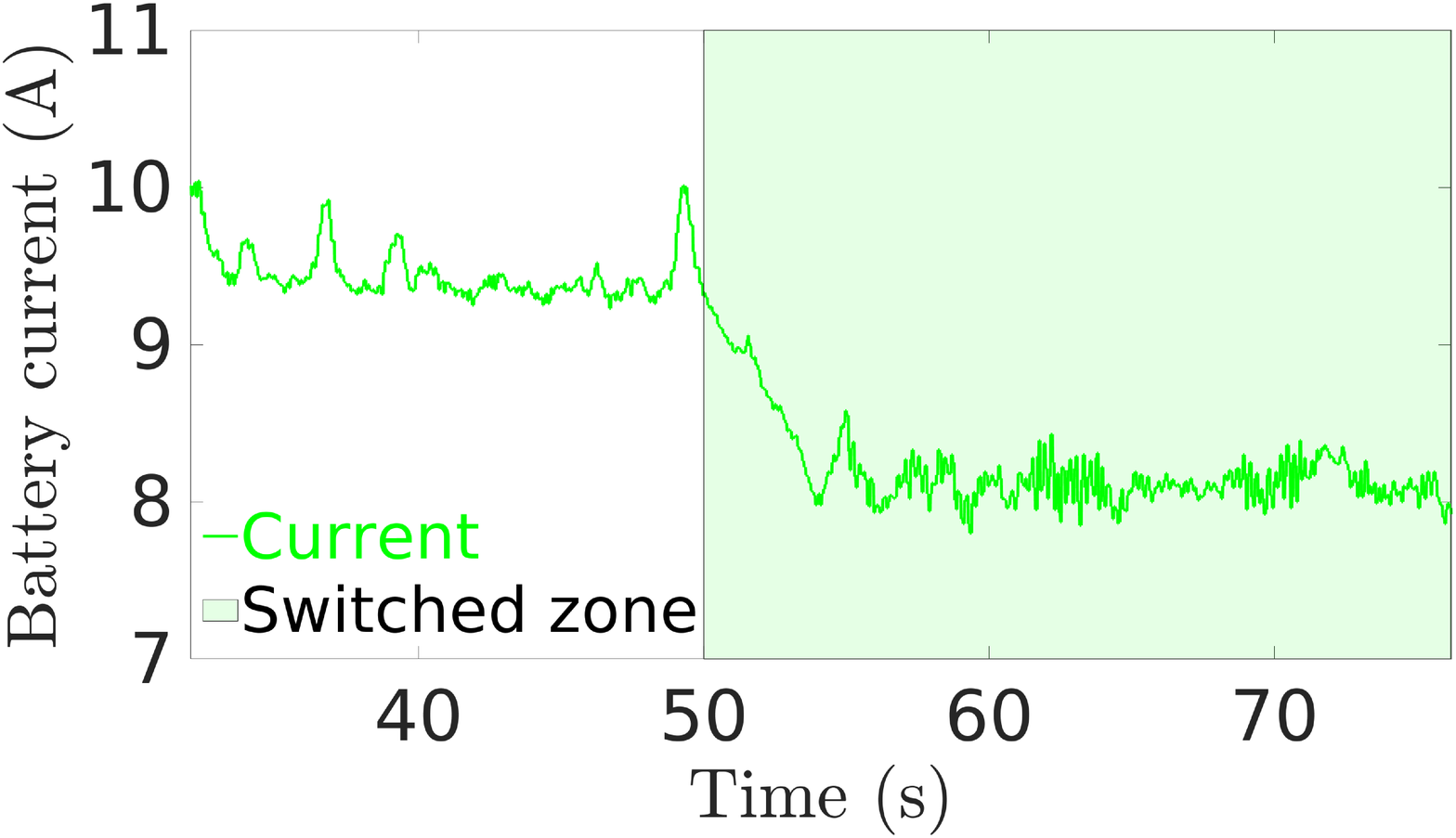}
		\label{fig_b_1}
	}
	\caption{The system performance for $\Delta_z/{\rm R} = 0.58$.}
	\label{ceiling1}	
\end{figure*}

One of the extreme cases, i.e., staying 1 cm below the ceiling, is tested in this implementation as can be seen in Fig. \ref{ceiling1}. The system still handles the ceiling effect in order not to be in permanent physical contact with the ceiling. Since the battery state is observed online in this implementation, it is explored that the current drawn from the battery can be decreased significantly during the flight in very close proximities (up to 15.8 \%). 

The average computation time of the NMPC is approximately 1.98 ms. When it is switched to the Force-NMPC, there is an increasing trend while the system approaches the ceiling (from 1.93 ms to 2 ms). The average computation time of the NMHE is around 3.35 ms. A similar rise is observed in the NMHE case, where the computation time changed between 3.29 ms to 3.38 ms while approaching the ceiling. 

\section{Discussion and Conclusion }\label{sec_conclusions}

In this work, we presented a force estimation based nonlinear MPC approach for operating quadrotors within close proximity to the surrounding. Our framework generates attitude angles and vertical force values that satisfy the dynamic behavior of the system together with the physical limits. Our algorithm is applicable in real-time and all the computations stay below 10 ms. We validated our approach experimentally in real-time using a small-scale quadrotor platform.

In the dual problem, the proposed estimation approach explores the defects (external forces, disturbances, unmodelled dynamics, and modeling mismatch) in the model leveraged within the controller. To this end, a suitable dynamical model for both free-flying and the interaction cases has been presented, together with an optimization framework for generating optimal motion reactions in close proximity. Our approach is agnostic to the information of the environment (e.g., distance to the ceiling), and hence eliminates the need for dedicated proximity or wind sensors and precise mathematical model.  

One main limitation of this work is to use the motion capture system, which provides precise motion information. The estimation-approach identifies the external forces online after each new pose measurement is available. For the in-situ inspection operation, we plan to adopt visual-inertial odometry methods for pose estimation in our future work.

There are several potential extensions of the proposed work. First of all, we intend to test our controller algorithm in more challenging environments with more complex interactions such as side walls and ceilings with stalactite-like structures. Second, the proposed approach is similar to the data-driven perspective. A ceiling effect model may be further explored with data collection. Third, the energy-aspects can be included in the cost function by mapping the current drawn from the batteries and the generated vertical forces to create a controller that is capable of prolonging flight duration and battery aging awareness \cite{Kumtepeli.2019}. This can allow to plan and execute efficient task-based trajectories to increase the flight envelope.

\section*{ACKNOWLEDGMENT}

The authors wish to thank for conducting the research work with support from the Energy Research Institute @ NTU (ERI@N) and NTU internal grant for the project on Large Vertical Take-Off and Landing (VTOL) Research Platform: Prototype development and demonstration (NTU internal funding).


\balance
\bibliography{References/bib,References/IEEEabrv}

\begin{thebibliography}{10}
\providecommand{\url}[1]{#1}
\csname url@rmstyle\endcsname
\providecommand{\newblock}{\relax}
\providecommand{\bibinfo}[2]{#2}
\providecommand\BIBentrySTDinterwordspacing{\spaceskip=0pt\relax}
\providecommand\BIBentryALTinterwordstretchfactor{4}
\providecommand\BIBentryALTinterwordspacing{\spaceskip=\fontdimen2\font plus
\BIBentryALTinterwordstretchfactor\fontdimen3\font minus
  \fontdimen4\font\relax}
\providecommand\BIBforeignlanguage[2]{{%
\expandafter\ifx\csname l@#1\endcsname\relax
\typeout{** WARNING: IEEEtran.bst: No hyphenation pattern has been}%
\typeout{** loaded for the language `#1'. Using the pattern for}%
\typeout{** the default language instead.}%
\else
\language=\csname l@#1\endcsname
\fi
#2}}

\bibitem{bircher2018receding}
A.~Bircher, M.~Kamel, K.~Alexis, \emph{et~al.}, ``Receding horizon path
  planning for 3d exploration and surface inspection,'' \emph{Autonomous
  Robots}, vol.~42, no.~2, pp. 291--306, 2018.

\bibitem{kocer2019inspection}
B.~B. Kocer, T.~Tjahjowidodo, M.~Pratama, \emph{et~al.},
  ``Inspection-while-flying: An autonomous contact-based nondestructive test
  using uav-tools,'' \emph{Automation in Construction}, vol. 106, p. 102895,
  2019.

\bibitem{kocer2018uav}
B.~B. Kocer, V.~Kumtepeli, T.~Tjahjowidodo, \emph{et~al.}, ``Uav control in
  close proximities-ceiling effect on battery lifetime,'' \emph{arXiv preprint
  arXiv:1812.11707}, 2018.

\bibitem{huber2013first}
F.~Huber, K.~Kondak, K.~Krieger, \emph{et~al.}, ``First analysis and
  experiments in aerial manipulation using fully actuated redundant robot
  arm,'' in \emph{Intelligent Robots and Systems (IROS), 2013 IEEE/RSJ
  International Conference on}.\hskip 1em plus 0.5em minus 0.4em\relax IEEE,
  2013, pp. 3452--3457.

\bibitem{conyers2018empirical}
S.~A. Conyers, M.~J. Rutherford, and K.~P. Valavanis, ``An empirical evaluation
  of ground effect for small-scale rotorcraft,'' in \emph{2018 IEEE
  International Conference on Robotics and Automation (ICRA)}.\hskip 1em plus
  0.5em minus 0.4em\relax IEEE, 2018, pp. 1244--1250.

\bibitem{bernard2018ground}
D.~D.~C. Bernard, M.~Giurato, F.~Riccardi, \emph{et~al.}, ``Ground effect
  analysis for a quadrotor platform,'' in \emph{Advances in Aerospace Guidance,
  Navigation and Control}.\hskip 1em plus 0.5em minus 0.4em\relax Springer,
  2018, pp. 351--367.

\bibitem{robinson2014computational}
D.~C. Robinson, H.~Chung, and K.~Ryan, ``Computational investigation of micro
  rotorcraft near-wall hovering aerodynamics,'' in \emph{Unmanned Aircraft
  Systems (ICUAS), 2014 International Conference on}.\hskip 1em plus 0.5em
  minus 0.4em\relax IEEE, 2014, pp. 1055--1063.

\bibitem{tavora2017feasibility}
B.~G. Tavora, ``Feasibility study of an aerial manipulator interacting with a
  vertical wall,'' Ph.D. dissertation, Monterey, California: Naval Postgraduate
  School, 2017.

\bibitem{hsiao_2018}
Y.~H. Hsiao and P.~Chirarattananon, ``Ceiling effects for surface locomotion of
  small rotorcraft,'' in \emph{Intelligent Robots and Systems (IROS), 2018
  IEEE/RSJ International Conference on}.\hskip 1em plus 0.5em minus 0.4em\relax
  IEEE, 2018.

\bibitem{conyers2018empiricalc}
S.~A. Conyers, M.~J. Rutherford, and K.~P. Valavanis, ``An empirical evaluation
  of ceiling effect for small-scale rotorcraft,'' in \emph{2018 International
  Conference on Unmanned Aircraft Systems (ICUAS)}.\hskip 1em plus 0.5em minus
  0.4em\relax IEEE, 2018, pp. 243--249.

\bibitem{yeo2015onboard}
D.~Yeo, N.~Sydney, D.~A. Paley, \emph{et~al.}, ``Onboard flow sensing for
  downwash detection and avoidance with a small quadrotor helicopter,'' in
  \emph{AIAA Guidance, Navigation, and Control Conference}, 2015, p. 1769.

\bibitem{yeo2016downwash}
D.~W. Yeo, N.~Sydney, D.~A. Paley, \emph{et~al.}, ``Downwash detection and
  avoidance with small quadrotor helicopters,'' \emph{Journal of Guidance,
  Control, and Dynamics}, vol.~40, no.~3, pp. 692--701, 2016.

\bibitem{sanchez2017experimental}
P.~Sanchez-Cuevas, G.~Heredia, and A.~Ollero, ``Experimental approach to the
  aerodynamic effects produced in multirotors flying close to obstacles,'' in
  \emph{Iberian Robotics conference}.\hskip 1em plus 0.5em minus 0.4em\relax
  Springer, 2017, pp. 742--752.

\bibitem{sanchez2017multirotor}
------, ``Multirotor uas for bridge inspection by contact using the ceiling
  effect,'' in \emph{Unmanned Aircraft Systems (ICUAS), 2017 International
  Conference on}.\hskip 1em plus 0.5em minus 0.4em\relax IEEE, 2017, pp.
  767--774.

\bibitem{delamare2018toward}
Q.~Delamare, P.~R. Giordano, and A.~Franchi, ``Toward aerial physical
  locomotion: The contact-fly-contact problem,'' \emph{IEEE Robotics and
  Automation Letters}, vol.~3, no.~3, pp. 1514--1521, 2018.

\bibitem{robinson2016ceiling}
D.~C. Robinson, H.~Chung, and K.~Ryan, ``Numerical investigation of a hovering
  micro rotor in close proximity to a ceiling plane,'' \emph{Journal of Fluids
  and Structures}, vol.~66, pp. 229 -- 253, 2016.

\bibitem{powers2013influence}
C.~Powers, D.~Mellinger, A.~Kushleyev, \emph{et~al.}, ``Influence of
  aerodynamics and proximity effects in quadrotor flight,'' in
  \emph{Experimental robotics}.\hskip 1em plus 0.5em minus 0.4em\relax
  Springer, 2013, pp. 289--302.

\bibitem{kocer2018centralized}
B.~B. Kocer, T.~Tjahjowidodo, and G.~G.~L. Seet, ``Centralized predictive
  ceiling interaction control of quadrotor vtol uav,'' \emph{Aerospace Science
  and Technology}, vol.~76, pp. 455--465, 2018.

\bibitem{svacha2017improving}
J.~Svacha, K.~Mohta, and V.~Kumar, ``Improving quadrotor trajectory tracking by
  compensating for aerodynamic effects,'' in \emph{Unmanned Aircraft Systems
  (ICUAS), 2017 International Conference on}.\hskip 1em plus 0.5em minus
  0.4em\relax IEEE, 2017, pp. 860--866.

\bibitem{bangura2017thrust}
M.~Bangura and R.~Mahony, ``Thrust control for multirotor aerial vehicles,''
  \emph{IEEE Transactions on Robotics}, vol.~33, no.~2, pp. 390--405, 2017.

\bibitem{tomic2018simultaneous}
T.~Tomi{\'c}, P.~Lutz, K.~Schmid, \emph{et~al.}, ``Simultaneous contact and
  aerodynamic force estimation (s-cafe) for aerial robots,'' \emph{arXiv
  preprint arXiv:1810.12908}, 2018.

\bibitem{yuksel2019aerial}
B.~Y{\"u}ksel, C.~Secchi, H.~H. B{\"u}lthoff, \emph{et~al.}, ``Aerial physical
  interaction via ida-pbc,'' \emph{The International Journal of Robotics
  Research}, 2019.

\bibitem{alexis2016aerial}
K.~Alexis, G.~Darivianakis, M.~Burri, \emph{et~al.}, ``Aerial robotic
  contact-based inspection: planning and control,'' \emph{Autonomous Robots},
  vol.~40, no.~4, pp. 631--655, 2016.

\bibitem{garimella2015towards}
G.~Garimella and M.~Kobilarov, ``Towards model-predictive control for aerial
  pick-and-place,'' in \emph{Robotics and Automation (ICRA), 2015 IEEE
  International Conference on}.\hskip 1em plus 0.5em minus 0.4em\relax IEEE,
  2015, pp. 4692--4697.

\bibitem{lunni_2017}
D.~Lunni, A.~Santamaria-Navarro, R.~Rossi, \emph{et~al.}, ``Nonlinear model
  predictive control for aerial manipulation,'' in \emph{2017 International
  Conference on Unmanned Aircraft Systems (ICUAS)}, June 2017, pp. 87--93.

\bibitem{seo2017aerial}
H.~Seo, S.~Kim, and H.~J. Kim, ``Aerial grasping of cylindrical object using
  visual servoing based on stochastic model predictive control,'' in
  \emph{Robotics and Automation (ICRA), 2017 IEEE International Conference
  on}.\hskip 1em plus 0.5em minus 0.4em\relax IEEE, 2017, pp. 6362--6368.

\bibitem{lee2010geometric}
T.~Lee, M.~Leok, and N.~H. McClamroch, ``Geometric tracking control of a
  quadrotor uav on se (3),'' in \emph{49th IEEE conference on decision and
  control (CDC)}.\hskip 1em plus 0.5em minus 0.4em\relax IEEE, 2010, pp.
  5420--5425.

\bibitem{quirynen2015autogenerating}
R.~Quirynen, M.~Vukov, M.~Zanon, \emph{et~al.}, ``Autogenerating microsecond
  solvers for nonlinear mpc: a tutorial using acado integrators,''
  \emph{Optimal Control Applications and Methods}, vol.~36, no.~5, pp.
  685--704, 2015.

\bibitem{kocer_tool}
B.~B. Kocer, T.~Tjahjowidodo, and G.~G.~L. Seet, ``Model predictive uav-tool
  interaction control enhanced by external forces,'' \emph{Mechatronics},
  vol.~58, pp. 47 -- 57, 2019.

\bibitem{kuhl_2011}
P.~Kuhl, M.~Diehl, T.~Kraus, \emph{et~al.}, ``A real-time algorithm for moving
  horizon state and parameter estimation,'' \emph{Computers \& Chemical
  Engineering}, vol.~35, no.~1, pp. 71 -- 83, 2011.

\bibitem{eren2017model}
U.~Eren, A.~Prach, B.~B. Ko{\c{c}}er, \emph{et~al.}, ``Model predictive control
  in aerospace systems: Current state and opportunities,'' \emph{Journal of
  Guidance, Control, and Dynamics}, vol.~40, no.~7, pp. 1541--1566, 2017.

\bibitem{houska2011acado}
\BIBentryALTinterwordspacing
B.~Houska, H.~J. Ferreau, and M.~Diehl, ``Acado toolkit—an open‐source
  framework for automatic control and dynamic optimization,'' \emph{Optimal
  Control Applications and Methods}, vol.~32, no.~3, pp. 298--312, 5 2011.
  [Online]. Available: \url{https://doi.org/10.1002/oca.939}
\BIBentrySTDinterwordspacing

\bibitem{ferreau2014qpoases}
H.~J. Ferreau, C.~Kirches, A.~Potschka, \emph{et~al.}, ``qpoases: A parametric
  active-set algorithm for quadratic programming,'' \emph{Mathematical
  Programming Computation}, vol.~6, no.~4, pp. 327--363, 2014.

\bibitem{Kumtepeli.2019}
V.~Kumtepeli, Y.~Zhao, M.~Naumann, \emph{et~al.}, ``Design and analysis of an
  aging-aware energy management system for islanded grids using mixed-integer
  quadratic programming,'' \emph{International Journal of Energy Research}, pp.
  1--18, 2019.

\end{thebibliography}
\bibliographystyle{References/IEEEtran}

\end{document}